\documentclass[letterpaper, 10 pt, conference]{ieeeconf}  

\IEEEoverridecommandlockouts                              

\usepackage{textcase}

\title{\LARGE \bf
Learning Actionable Manipulation Recovery \\ via Counterfactual Failure Synthesis}

\author{
Dayou Li$^{1}$, Jiuzhou Lei$^{1}$, Hao Wang$^{1}$, Lulin Liu$^{1,2}$, Yunhao Yang$^{3}$, Zihan Wang$^{4}$, Bangya Liu$^{5}$ \\ Minghui Zheng$^{1}$, Zhiwen Fan$^{1}$ \\
$^{1}$Texas A\&M University, $^{2}$University of Minnesota, $^{3}$University of Texas at Austin \\ $^{4}$Abaka AI, $^{5}$University of Wisconsin-Madison \\
\url{https://dream2fix.github.io/}
}
\usepackage{ragged2e}
\usepackage{array}
\usepackage{tabularx}
\usepackage{makecell}
\usepackage[most]{tcolorbox}
\usepackage{cite}
\usepackage{amsmath}
\usepackage{amssymb}
\usepackage{booktabs}   
\usepackage{multirow}   
\usepackage{graphicx} 
\usepackage{xcolor}

\usepackage{bbm}  

\DeclareMathOperator{\median}{median}

\makeatletter
\def\fnum@table{TABLE~\thetable}

\long\def\@makecaption#1#2{%
  \vskip\abovecaptionskip
  \footnotesize
  \sbox\@tempboxa{#1:\hspace{0.5em}#2}%
  \ifdim \wd\@tempboxa >\hsize
    #1:\hspace{0.5em}#2\par
  \else
    \hbox to\hsize{\hfil\box\@tempboxa\hfil}%
  \fi
  \vskip\belowcaptionskip
}
\makeatother

\usepackage{url}
\usepackage{caption}
\begin{document}

\maketitle
\thispagestyle{empty}
\pagestyle{empty}

\begin{abstract}
While recent foundation models have significantly advanced robotic manipulation, these systems still struggle to autonomously recover from execution errors. Current failure-learning paradigms rely on either costly and unsafe real-world data collection or simulator-based perturbations, which introduce a severe sim-to-real gap. Furthermore, existing visual analyzers predominantly output coarse, binary diagnoses rather than the executable, trajectory-level corrections required for actual recovery. To bridge the gap between failure diagnosis and actionable recovery, we introduce Dream2Fix, a framework that synthesizes photorealistic, counterfactual failure rollouts directly from successful real-world demonstrations. By perturbing actions within a generative world model, Dream2Fix creates paired failure-correction data without relying on simulators. To ensure the generated data is physically viable for robot learning, we implement a structured verification mechanism that strictly filters rollouts for task validity, visual coherence, and kinematic safety. This engine produces a high-fidelity dataset of over 120k paired samples. Using this dataset, we fine-tune a vision-language model to jointly predict failure types and precise recovery trajectories, mapping visual anomalies directly to corrective actions. Extensive real-world robotic experiments show our approach achieves state-of-the-art correction accuracy, improving from 19.7\% to 81.3\% over prior baselines, and successfully enables zero-shot closed-loop failure recovery in physical deployments.

\end{abstract}

\section{Introduction}
Recent foundation models, including vision-language models (VLMs) and vision-language-action models (VLAs), have introduced open-vocabulary perception and strong generalization to robotic manipulation~\cite{driess2023palme,zitkovich2023rt2,bousmalis2023robocat,ji2025robobrain,yuan2024robopoint}. Despite these advances, such models struggle to autonomously recover from real-world execution errors. While they excel at nominal task execution, they lack the causal understanding of failure dynamics required to synthesize executable corrective actions. Autonomous recovery necessitates not just detecting an anomaly, but reasoning about its root cause and translating that diagnosis into precise trajectory corrections. Developing this capability is severely bottlenecked by data; real-world failures rarely occur with paired, dense correction annotations, and intentionally collecting physical failure data across diverse tasks is both costly and hazardous to hardware.

Prior efforts to endow foundation models with failure-recovery capabilities largely follow two paradigms. The first utilizes reward signals or preference annotations to refine policies~\cite{lee2026roboreward,peng2026reworld,intelligence2025pi}. However, these methods frequently rely on video-level temporal heuristics that yield low-fidelity supervision, or depend on non-scalable human-in-the-loop feedback. The second paradigm scales data collection through procedural perturbations in simulation, training VLMs to output failure taxonomies~\cite{duan2025aha}. While scalable, simulation-based generation suffers from a pronounced sim-to-real gap. Furthermore, these visual analyzers typically provide coarse, binary diagnoses rather than the low-level, executable trajectory corrections necessary for closed-loop recovery. Consequently, a persistent gap remains between high-level failure detection and actionable recovery (as shown in Tab.~\ref{tab:aha_vqa_lvis}).

\begin{figure}[t]
  \centering
  \includegraphics[width=1\linewidth]{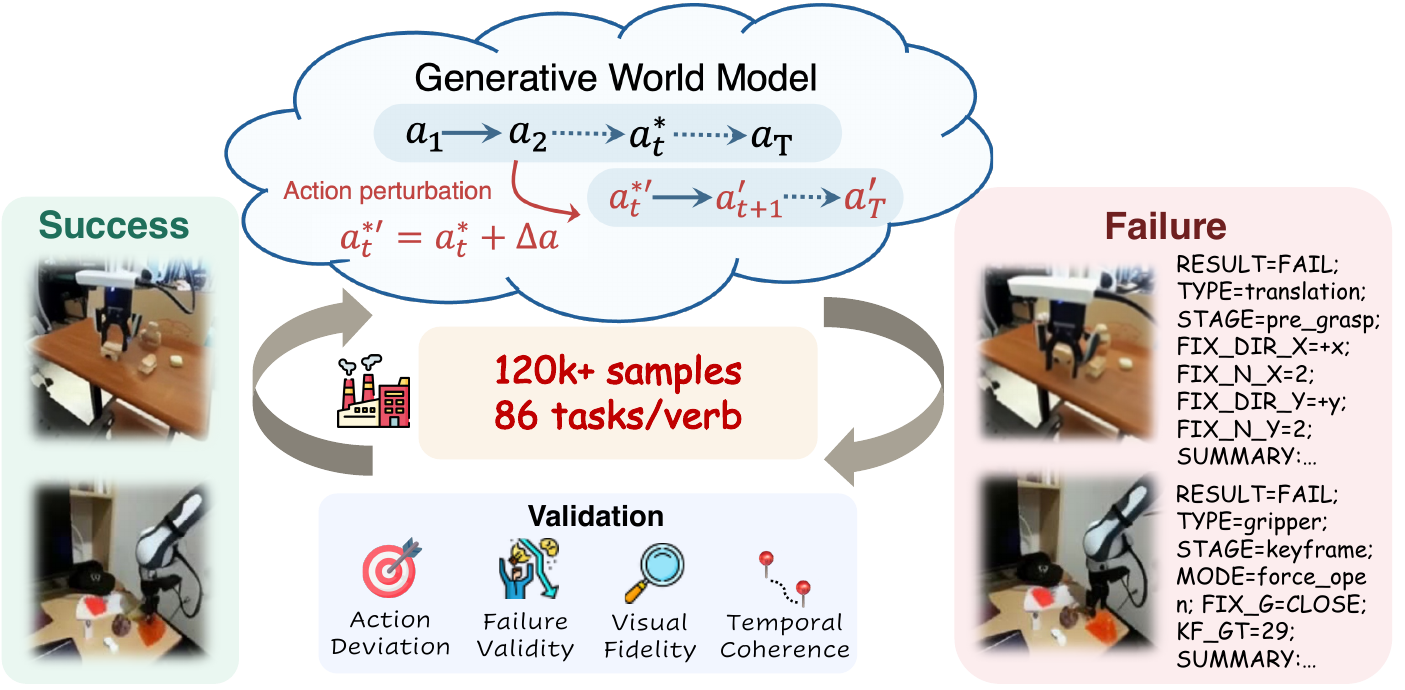} 
  \caption{Dream2Fix is a data generation pipeline that synthesizes large-scale, photorealistic failure rollouts with paired corrections from successful demonstrations and curates them with physical and visual verification.
  }
  \label{fig:teaser}
  \vspace{-0.3cm}
\end{figure}

To address these limitations, we introduce Dream2Fix, a data generation framework that synthesizes large-scale, photorealistic, and physically plausible counterfactual failure rollouts directly from successful real-world demonstrations (Fig.~\ref{fig:teaser}). By injecting targeted action perturbations (e.g., end-effector pose offsets) into a generative world model, Dream2Fix produces diverse failure videos paired with deterministic trajectory-level corrections. Because generative models can introduce visual or physical artifacts, we implement a structured verification mechanism. This verifier strictly filters synthesized samples across three dimensions: task failure validity, visual coherence, and kinematic safety. This ensures that retained rollouts accurately reflect the intended failure mode and yield executable corrective motions within robot constraints.

Using this engine, we curate a high-fidelity dataset of over 120,000 paired failure-correction samples, simultaneously mitigating the sim-to-real gap and providing denser supervision than prior datasets. We formulate failure recovery as a spatial reasoning task and fine-tune a VLM on this dataset to jointly predict failure diagnoses and precise trajectory corrections, directly mapping visual anomalies to recovery actions. Extensive evaluations demonstrate that our approach significantly improves correction accuracy over established baselines, from 19.7\% to 81.3\%, and achieves a 46\% zero-shot recovery rate during real-robot closed-loop deployment.

To summarize, our main contributions are:
\begin{itemize}
    \item We propose Dream2Fix, a data generation pipeline that leverages a generative world model to synthesize photorealistic robotic failure cases from successful demonstrations, governed by a rigorous physical and visual verifier.
    \item We introduce a failure-aware VLM that bridges diagnosis and actionable recovery by jointly predicting structured failure causes and precise, trajectory-level corrections.
    \item We establish a comprehensive evaluation framework for failure reasoning, demonstrating significant performance gains in both offline benchmarks and zero-shot, closed-loop real-robot recovery.
\end{itemize}

\begin{figure*}[t]
  \centering
  \includegraphics[width=1\linewidth]{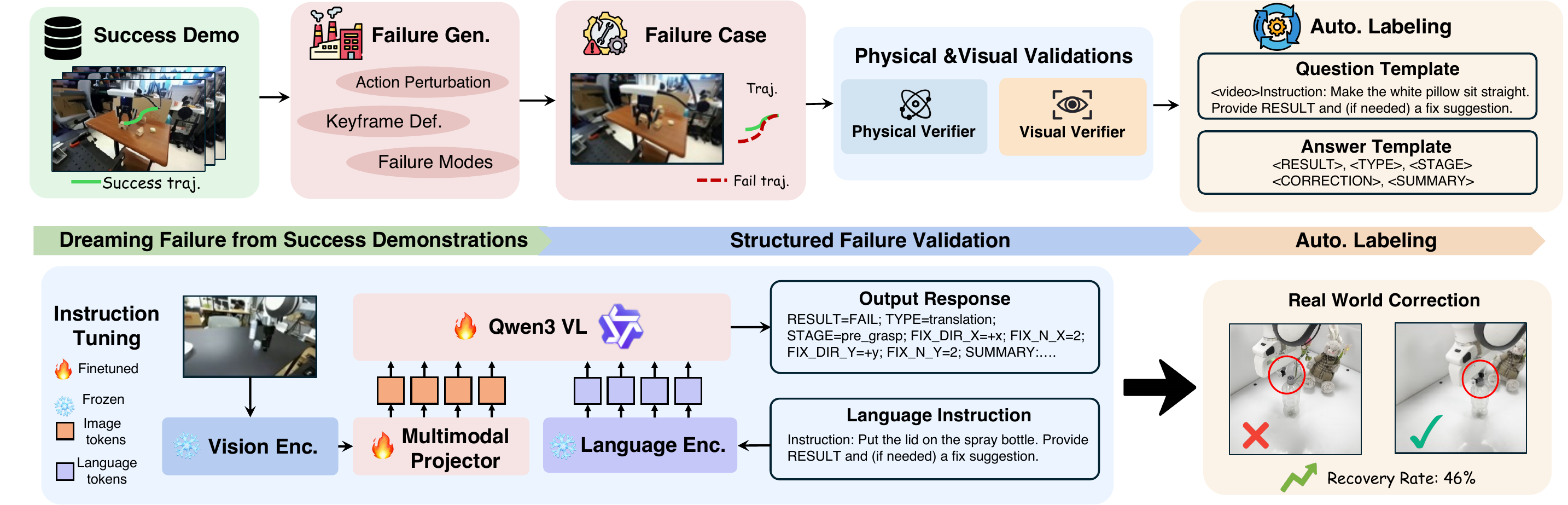} 
  \caption{\textbf{Overview of Dream2Fix pipeline}. Dream2Fix generates diverse failure cases from successful demonstrations via keyframe-level action perturbations, then validates and curates them with physical and visual verifiers. The verified rollouts are auto-labeled into a structured schema to instruction-tune a VLM that predicts actionable corrections for real-world recovery.
  }
  \label{fig:pipeline}
  \vspace{-0.3cm}
\end{figure*}
\section{Related Works}

\subsection{Failure Detection in Robotic Manipulation}
Failure in robotic manipulation is a common and critical challenge, and recent work has explored both detecting failure events and reasoning about their underlying causes. Gu et al.~\cite{gu2025safe} analyze VLA feature representations to predict task failure likelihood, but do not provide corrective actions. 
Pacaud et al.~\cite{pacaud2025guardian} combine a simulator and limited real-robot perturbations to train a VLM for failure reasoning, but the real-world supervision is weak and misses execution-level failures. Liu et al.~\cite{liu2023reflect} collect failures from both simulation and real robots, but the dataset is small and limited in task and environment diversity. Qi et al.~\cite{qi2026selfrefining} use iterative self-refinement for failure detection and reasoning, but still rely on limited annotations and lack structured corrective supervision. Overall, prior works provide coarse failure reasoning but lacks actionable trajectory-level corrections, and remains constrained by unrealistic simulations or small real-world datasets, which makes it hard to generalize across diverse manipulation settings.

\subsection{Data Generation in Robotics}
A growing number of systems focus on scaling the collection of robotic data for training manipulation policies. Simulation-based pipelines have been widely adopted, such as MimicGen~\cite{mandlekar2023mimicgen}, SAGE~\cite{xia2026sagescalableagentic3d}, and RoboGen~\cite{wang2024robogen}, which enable large-scale data collection and offer fine-grained control over task and environment. However, since they are conducted entirely in simulation, the resulting data often lacks visual realism, which brings large sim2real gap.
Recent progress in world models suggests an alternative direction for generating realistic, action-conditioned data at scale~\cite{pmlr_v235_bruce24a,pmlr_v235_zhou24f,guo2025ctrlworld,hafner2025diverseworldmodels,ye2025anchordream}, though ensuring physical plausibility remains challenging. These trends motivate leveraging world models to synthesize realistic failure data while incorporating explicit validity assessments to mitigate unrealistic generations.

\subsection{Robotic Vision-Language Models}

Recent robotic VLMs improve spatial grounding for manipulation, but they are mostly designed for perception and planning rather than failure-driven correction. SpatialBot~\cite{cai2025spatialbot} uses RGB-D supervision and hierarchical spatial Q\&A to strengthen depth-aware understanding, but it does not learn structured recovery targets from failures. RoboRefer~\cite{zhou2025roborefer} resolves complex referential expressions via multi-stage spatial reasoning, but it does not support failure detection or recovery. RoboPoint~\cite{yuan2025robopoint} predicts dense affordance maps for where actions are feasible, yet it offers limited trajectory-level guidance for correcting execution errors.

\section{Method}

\subsection{Problem Statement}
We consider the task of diagnosing and correcting robotic manipulation failures from visual observations. Formally, given an input video sequence $\mathcal{I} = \{I_0, I_1, \dots, I_T\}$, where $I_t \in \mathbb{R}^{H \times W \times 3}$ is an RGB frame at timestep $t$, the objective is to predict three outputs:

\begin{itemize}
    \item \textbf{Result Classification:} a categorical label $y_{\text{result}} \in \{\text{success}, \text{fail}\}$ indicating the outcome of the manipulation attempt.
    
    \item \textbf{Failure Type:} a categorical label $y_{\text{fail}} \in \mathcal{F}$, where $\mathcal{F} = \{\text{translation}, \text{weak\_close}, \text{force\_open}, \text{delay\_close}\}$ defines a set of failure modes. 
    
    \item \textbf{Actionable Correction} a structured textual output $y_{\text{fix}} \in \mathcal{T}$ consisting of correction directives and a natural language summary. The directive fields specify how to correct the failure, including gripper actions at the keyframe or spatial adjustments.

\end{itemize}

We formulate failure recovery as structured prediction and fine-tune a VLM to jointly output success/failure, failure types, and trajectory-level recovery under a fixed schema. This structured supervision makes corrections interpretable and directly usable by robot controllers (as shown in Fig.~\ref{fig:pipeline}).

\subsection{Photorealistic Failure Generation}\label{sec:dataset}
\subsubsection{Failure Type Definition and Rollout Collection}
To construct a dataset of diverse and interpretable failure cases, we define four representative failure types in robotic manipulation: \text{delay\_close}, \text{weak\_close}, \text{force\_open}, and \text{translation}. These failure modes reflect common execution errors related to gripper control and spatial accuracy. Specifically:

\begin{itemize}
    \item \text{delay\_close}: the gripper closes with a significant delay relative to the intended grasping moment, resulting in a late or missed closure.
    
    \item \text{weak\_close}: the gripper closes too softly or with incomplete force, which often leads to unstable contact and potential object drop.
    
    \item \text{force\_open}: the gripper fails to close entirely at the designated keyframes, remaining open due to prediction errors by robotic policy. 
    
    \item \text{translation}: perturbations in the end-effector's $x$-$y$ states during the approach and contact phase, often cause position misalignment or grasp failure
\end{itemize}

These types were chosen based on the subset of failure modes that can be stably induced and reproduced using current robotic world models. We adopt \text{Ctrl-World}~\cite{guo2025ctrlworld} as the base generator. Unlike prior models that take in task descriptions, Ctrl-World is directly conditioned on robot actions, which we perturb to synthesize failure cases.

We define the \text{keyframe} as the timestep at which the gripper transitions between open and close states. For each rollout, we perturb actions over a local temporal window around the keyframes, thereby yielding more continuous rollouts and a higher probability of the desired failure.
For instance, we add temporal offsets to delay gripper closure (\text{delay\_close}), reduce the closure magnitude for \text{weak\_close}, or invert the gripper action for \text{force\_open}. For \text{translation} failures, we introduce Gaussian noise to the end-effector’s $x$ and $y$ states near. These perturbations are minimal but sufficient to produce visible and meaningful deviations in the final rollout. Each video is annotated with a result label, a failure type, and a correction in structured text format. This strategy enables scalable and consistent synthesis of realistic failure data without any real world cost.

\subsubsection{Structured Failure Validation}
\label{sec:verifiers}
To ensure the physical plausibility and visual quality of the synthesized failures, we employ four verifiers to screen generated samples along four validation dimensions: task failure validity, visual quality (including temporal coherence), dynamics consistency, and kinematic safety. A sample is retained only if it satisfies all verifier conditions. We next describe the design and implementation of each verifier.
A generated sample is retained only if it satisfies the corresponding validation criteria. In the following, we elaborate each verifier.

\newcolumntype{Y}{>{\centering\arraybackslash}X}
\newcolumntype{L}{>{\RaggedRight\arraybackslash}X}
\newcolumntype{S}{>{\centering\arraybackslash}p{1.55cm}}

\begin{table*}[t]
\centering
\caption{\textbf{Comparison to existing datasets.} We compare Dream2Fix with AHA~\cite{duan2025aha} and REFLECT~\cite{liu2023reflect} in terms of dataset scale, input modalities, and label style. Dream2Fix pairs synthesized failure clips with structured labels that specify failure type and stage, together with executable corrections, enabling vision-language models to produce grounded failure diagnoses and executable corrections.}
\label{tab:aha_vqa_lvis}

\small
\setlength{\tabcolsep}{5pt}
\renewcommand{\arraystretch}{1.05}

\resizebox{0.9\textwidth}{!}{%
\begin{tabularx}{\textwidth}{@{}SLLL@{}}
\toprule
\textbf{Source} &
\multicolumn{1}{Y}{\textbf{Dream2Fix}} &
\multicolumn{1}{Y}{\textbf{AHA \cite{duan2025aha}}} &
\multicolumn{1}{Y}{\textbf{REFLECT \cite{liu2023reflect}}} \\
\midrule

& \multicolumn{1}{Y}{\makebox[\linewidth][c]{\includegraphics[width=0.28\textwidth]{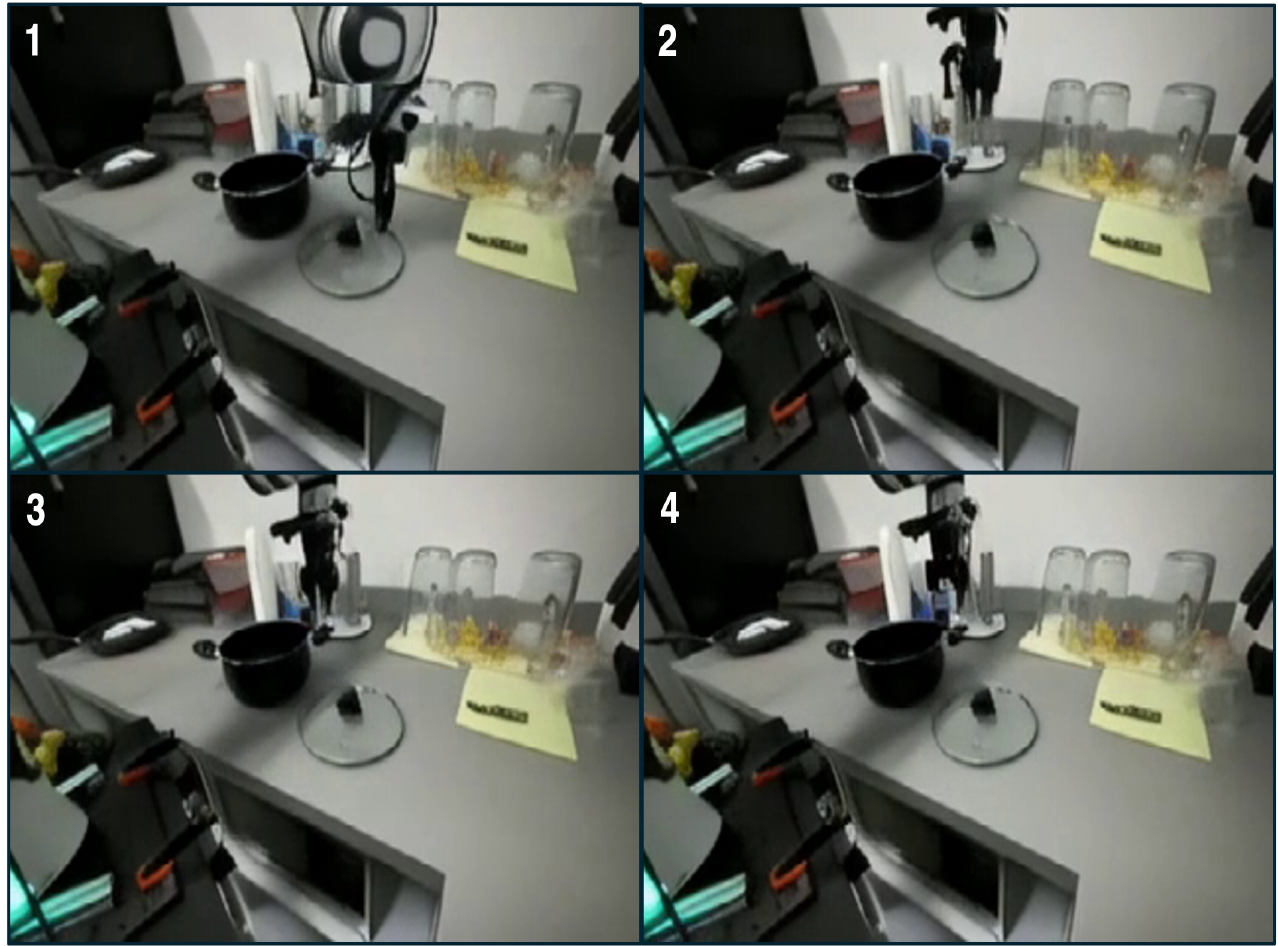}}}
& \multicolumn{1}{Y}{\makebox[\linewidth][c]{\includegraphics[width=0.28\textwidth]{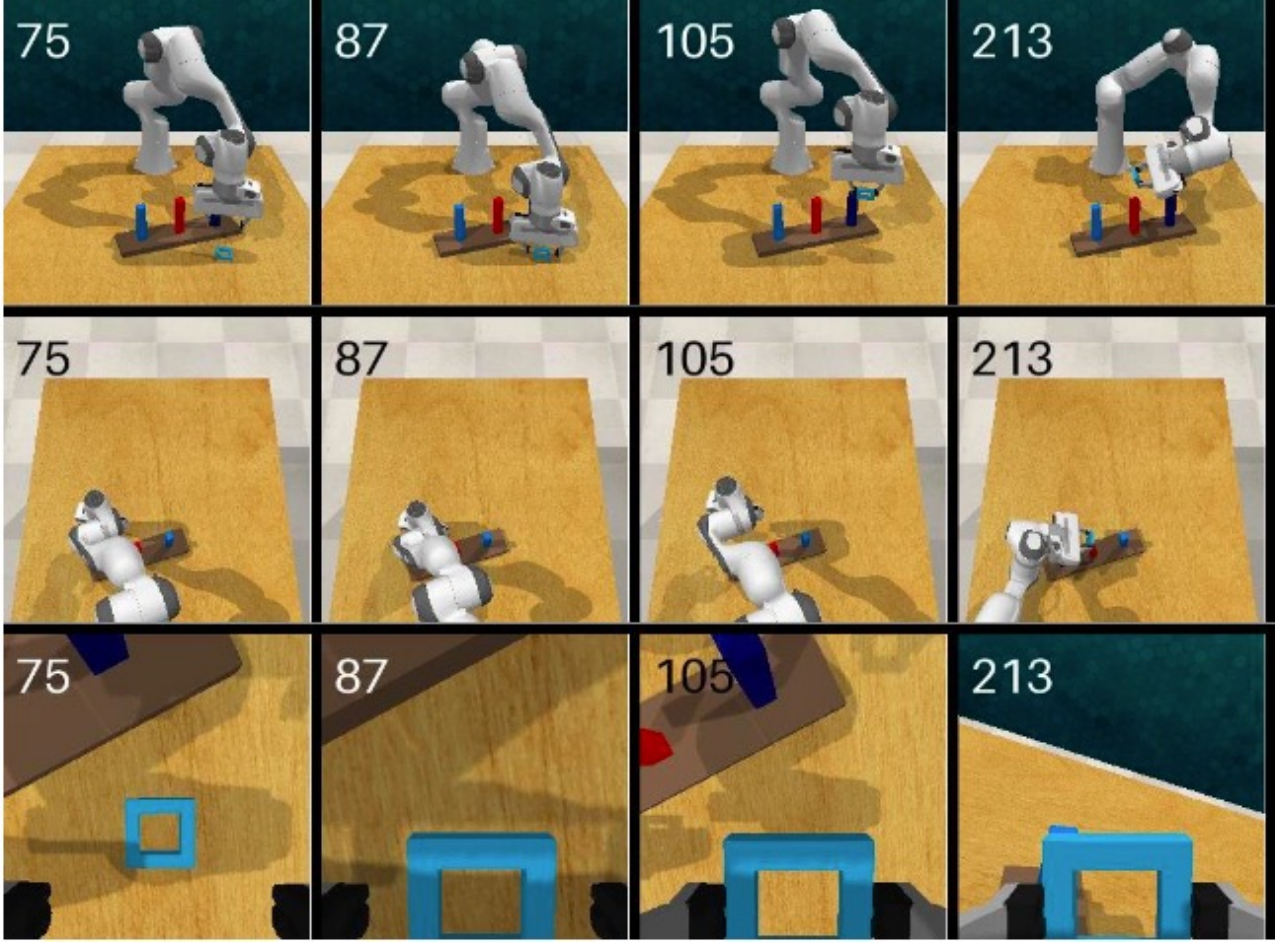}}}
& \multicolumn{1}{Y}{\makebox[\linewidth][c]{\includegraphics[width=0.28\textwidth]{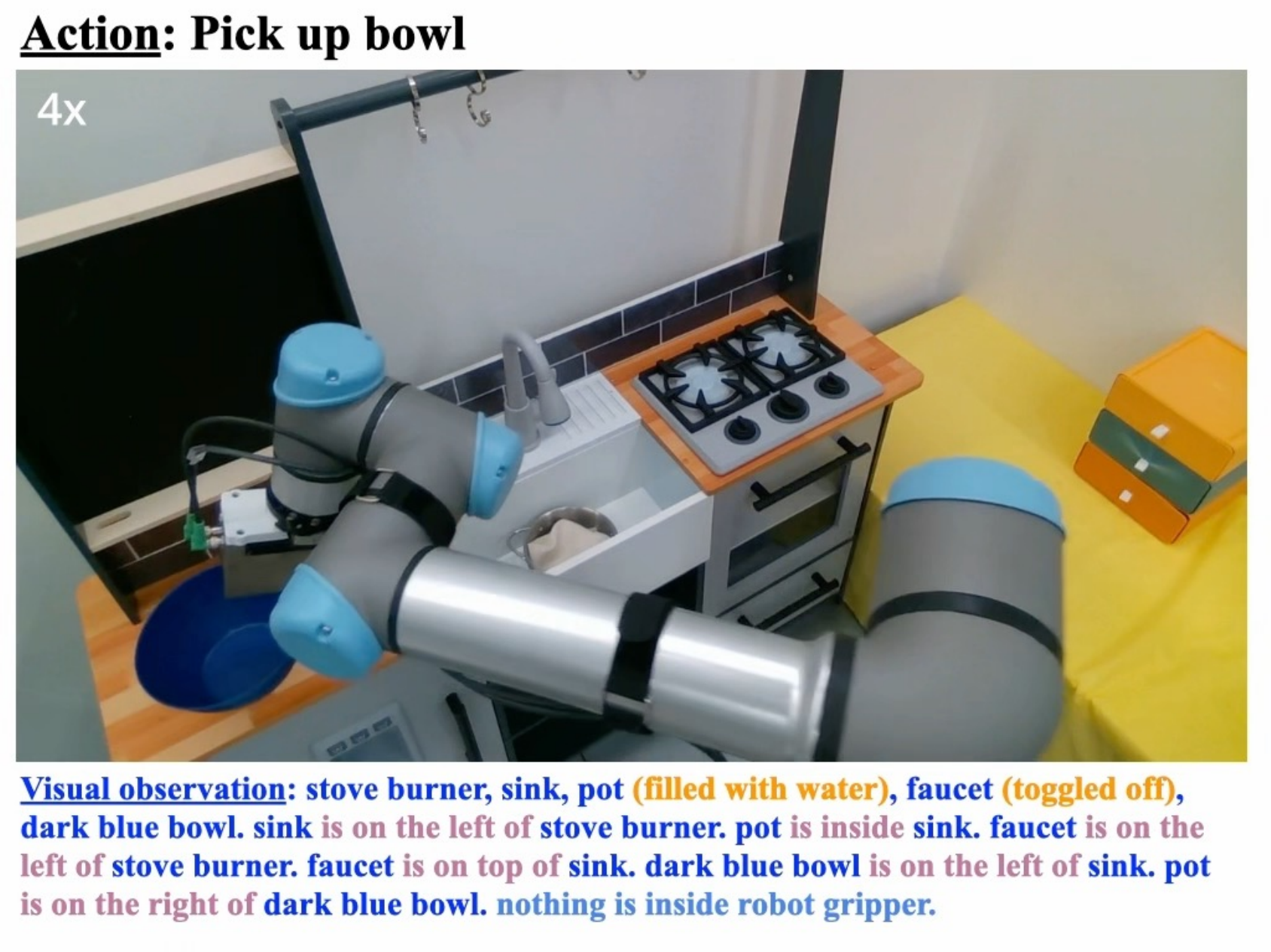}}} \\
\midrule

\textbf{Quantity}
& \multicolumn{1}{Y}{120K}
& \multicolumn{1}{Y}{49K}
& \multicolumn{1}{Y}{130} \\
\midrule

\textbf{Input} &
Instruction: Put the lid on the pot. Provide RESULT and (if needed) a fix suggestion.
&
For the given sub-tasks, first determine if it has succeeded by choosing from [``yes'', ``no''] and then explain the reason why the current sub-tasks have failed.
&
RGB-D video frames, audio, robot proprioception \\
\midrule

\textbf{Label} &
RESULT=FAIL; TYPE=translation; STAGE=pre\_grasp; FIX\_DIR\_X=-x; FIX\_N\_X=2; FIX\_DIR\_Y=+y; FIX\_N\_Y=3; The execution failed due to a translation misalignment before grasping. To fix it, nudge the end-effector in -x for 2 steps and in +y for 3 steps in the keyframe.
&
No, the robot gripper rotated with an incorrect roll angle.
&
At 03:00, the robot failed to properly place the pot filled with water on the stove burner. The failure was caused by the robot dropping something on the ground at 02:43, which might have been the pot or its contents. Failure timestep: 02:43 \\
\bottomrule
\end{tabularx}%
}
\vspace{-0.5cm}
\end{table*}

\noindent\textbf{VLM Verifier.}
We design a VLM-based verifier to jointly assess whether a generated failure clip (i) truly manifests the intended failure and (ii) maintains adequate visual quality without synthesis artifacts or distortions. Concretely, we apply Qwen3-VL-8B as the base, together with the task instruction, the reference success clip, and the generated failure clip, to provide two judgments: task failure validity (whether the generated execution should be considered a failure compared to the reference) and visual quality (whether the video contains severe hallucinations). We retain a synthesized sample only when it is judged as a valid failure and passes the visual-quality criterion.


\noindent\textbf{Inverse Dynamics Model (IDM) Verifier.}
We first train an IDM on successful demonstrations to capture the domain-specific action--state relationship. We instantiate the IDM with a ResNet-50~\cite{he2016deep} visual backbone followed by an MLP head that regresses the state difference between two frames.
Given two frames $I_t$ and $I_{t+d}$ from a generated rollout, the IDM $f_{\theta}$ infers the robotic end-effector's state difference implied by the visual transition:
\begin{equation}
\widehat{\Delta s}_{t,d}=f_{\theta}\!\left(I_t, I_{t+d}\right)
\label{eq:idm_pred}
\end{equation}
Each generated clip is synthesized conditioned on the recorded state $s_t$ and action sequence, from which we compute the reference state difference
\begin{equation}
\Delta s_{t,d}= s_{t+d}-s_t 
\label{eq:idm_ref}
\end{equation}
We measure the discrepancy between the visually inferred transition and the conditioning signal via
\begin{equation}
e_{t,d}=\left\lVert \widehat{\Delta s}_{t,d}-\Delta s_{t,d}\right\rVert_2 
\label{eq:idm_err}
\end{equation}
and accept a synthesized sample only if $e_{t,d}\le \tau$, where $\tau$ is calibrated on successful reference trajectories.

\noindent\textbf{Joint Pose Verifier.}
We discard kinematically unsafe rollouts by evaluating joint limit and motion smoothness constraints in joint space.
We train a joint pose predictor based on ResNet-50~\cite{he2016deep} to estimate per-frame joint angles $q_t\in\mathbb{R}^{7}$ from the generated video, and compute joint velocities $\dot{q}_t$ and accelerations $\ddot{q}_t$ using finite differences.
Thresholds are calibrated on successful reference trajectories.
We reject a synthesized clip if there exists a time step $t$ and joint index $j$ such that the predicted joint angle violates the joint limits, or the joint velocity exceeds a threshold, or the joint acceleration exceeds a threshold:
\[
\exists\, t,j:\quad
q_{t,j}\notin[q_j^{\min},q_j^{\max}]
\ \ \text{or}\ \ 
|\dot{q}_{t,j}|>\tau_v
\ \ \text{or}\ \ 
|\ddot{q}_{t,j}|>\tau_a .
\]
This verifier removes rollouts that appear visually plausible but imply unsafe or physically implausible joint dynamics.



\noindent\textbf{Point-Tracking Verifier.}
We assess temporal coherence via point tracking~\cite{karaev24cotracker3}. Given $M$ tracked points (uniformly sampled on a $10\times10$ grid in the initial frame), we obtain 2D trajectories $\{p_t^i\}_{t=1}^{T}$ in pixel space with visibility masks $\{m_t^i\}_{t=1}^{T}$. We compute four complementary metrics, including motion smoothness $S_{\text{smooth}}$, visibility stability $S_{\text{vis}}$, local topology stability $S_{\text{topo}}$, and global continuity $S_{\text{global}}$, and combine them into a total score $S_{\text{pt}}$ for dataset curation. We clip all scores to the range $[0,1]$. We additionally discard clips with insufficient tracking confidence. Let $a$ denote the per-track second-order motion magnitude (discrete acceleration) computed from $\{p_t^i\}$, and let $r_{\text{spike}}$ be the fraction of frames exhibiting spike-like transients. We define:
\begin{equation}
S_{\text{smooth}}=\exp\!\left(-\frac{Q_{0.95}(a)}{\tau_{\text{acc}}}\right)\cdot(1-r_{\text{spike}})
\end{equation}
\begin{equation}
S_{\text{vis}}=1-\median_{i}\!\left(\frac{1}{T-1}\sum_{t}\mathbf{1}\!\left[m_{t+1}^i \neq m_t^i\right]\right)
\end{equation}
We form KNN edges in the first frame and track their pairwise distances over time. For an edge $(i,j)$, let $d_t^{ij}$ denote the distance at time $t$. We define:
\begin{equation}
u_t^{ij}=\frac{\left|d_t^{ij}-d_0^{ij}\right|}{d_0^{ij}+\epsilon},
\qquad
S_{\text{topo}}=\exp\!\left(-\frac{\median(u)}{\tau_{\text{topo}}}\right)
\end{equation}
We fit an affine transform between adjacent frames and compute per-frame alignment errors (rmse) and jitter, defined as the temporal fluctuation of the estimated global affine parameters across consecutive frame pairs. Thus, we define:
\begin{equation}
\begin{aligned}
S_{\text{global}}
&=0.7\,\exp\!\left(-\frac{Q_{0.9}(\mathrm{rmse})}{\tau_{\mathrm{rmse}}}\right)\\
&\qquad\ \ +0.3\,\exp\!\left(-\frac{Q_{0.9}(\mathrm{jitter})}{\tau_{\mathrm{jitter}}}\right)
\end{aligned}
\end{equation}
Finally, we compute the total score:
\begin{equation}
\begin{aligned}
S_{\text{pt}}
&=w_1S_{\text{smooth}}+w_2S_{\text{vis}}\\
&\qquad\ \ +w_3S_{\text{topo}}+w_4S_{\text{global}}.
\end{aligned}
\end{equation}

\subsubsection{Fix Suggestion Generation}
For each verified failure rollout, we generate an actionable recovery as a structured text label that can be directly parsed by downstream modules.
Specifically, the actionable recovery follows a fixed key--value schema, followed by a short natural language summary.
The key--value fields specify actionable recovery in the trajectory, while the summary briefly describes the failure mode and the intended correction.
Actionable recoveries are generated deterministically from the injected perturbation parameters used to generate the failure from success demonstrations, yielding consistent self-paired corrective supervision.

For gripper failures (delay\_close, weak\_close, force\_open), the actionable recovery specifies the corrected gripper command and a temporal anchor indicating when the gripper state change should be executed. This is necessary because gripper failures may prevent the state change, making the keyframe ambiguous in the failed rollout. For translation failures, the actionable recovery specifies an axis-aligned correction direction and a discrete step count along each axis.

We avoid free-form recovery generation during dataset construction and instead adopt a predefined schema to keep the output space stable and unambiguous. Each dataset sample contains (i) a result label (success or fail), (ii) a failure mode, and (iii) an actionable recovery in a structured key--value format, optionally accompanied by a short natural language summary. The structured fields can be mapped to corrective control commands for the robot controllers.

\subsection{Learn from Failure}
\subsubsection{VLM Finetuning}
We follow a standard visual instruction-tuning paradigm to adapt Qwen3-VL-8B-Instruct to our failure reasoning and correction task. The model consists of a vision encoder, a multimodal projector that maps visual features into the language embedding space, and a transformer-based language model that generates text outputs autoregressively. Given a single-view manipulation video and the task instruction, the model is trained to output a structured response that includes the result label (\text{success}/\text{fail}), the failure type (\text{delay\_close}, \text{weak\_close}, \text{force\_open}, \text{translation}), and an actionable correction in a key--value format with a natural language summary. We fine-tune the VLM on our Dream2Fix dataset constructed in Sec.~\ref{sec:dataset}, and name the model as Dream2Fix-VLM. All components are initialized from the pre-trained checkpoint. During fine-tuning, we update the language model parameters and the multimodal projector parameters, while keeping other components frozen.

\subsubsection{Closed-Loop Recovery for VLA}
Dream2Fix enables real-world closed-loop recovery by grounding the VLM outputs into executable robot control primitives, thereby improving downstream VLA performance. We execute a base VLA policy on the real robot and record the execution trajectory. 
The record will first be downsampled and processed by Dream2Fix-VLM with the structured prompt, in order to reason about the failure. A deterministic mapping module is designed to parse the action recovery into robot control primitives, which are executed to correct the failed trajectory.

In practice, we select the keyframe and decide when to invoke the VLM using an event-aligned rule with a fixed action budget. If no closing attempt occurs within the budget, we terminate the rollout and extract a fixed-length clip from the final portion of the trajectory, then invoke the VLM to handle cases where the policy fails to initiate the gripper action.
To execute the predicted actionable recovery, the mapping module applies a deterministic conversion from structured fields to control primitives. For translation failures, we convert the predicted axis-aligned adjustment into a corresponding translation delta in the robot control space. For gripper failures, including missing, delayed, or weak closing, we issue an explicit gripper-close command at the predicted time and, when necessary, re-close or increase closing strength to ensure a firm grasp. After applying recovery, we replay the edited trajectory, record videos together with per-frame states and actions, and add recovered rollouts to the training database.

\begin{figure}[t]
  \centering
  \includegraphics[width=0.8\linewidth]{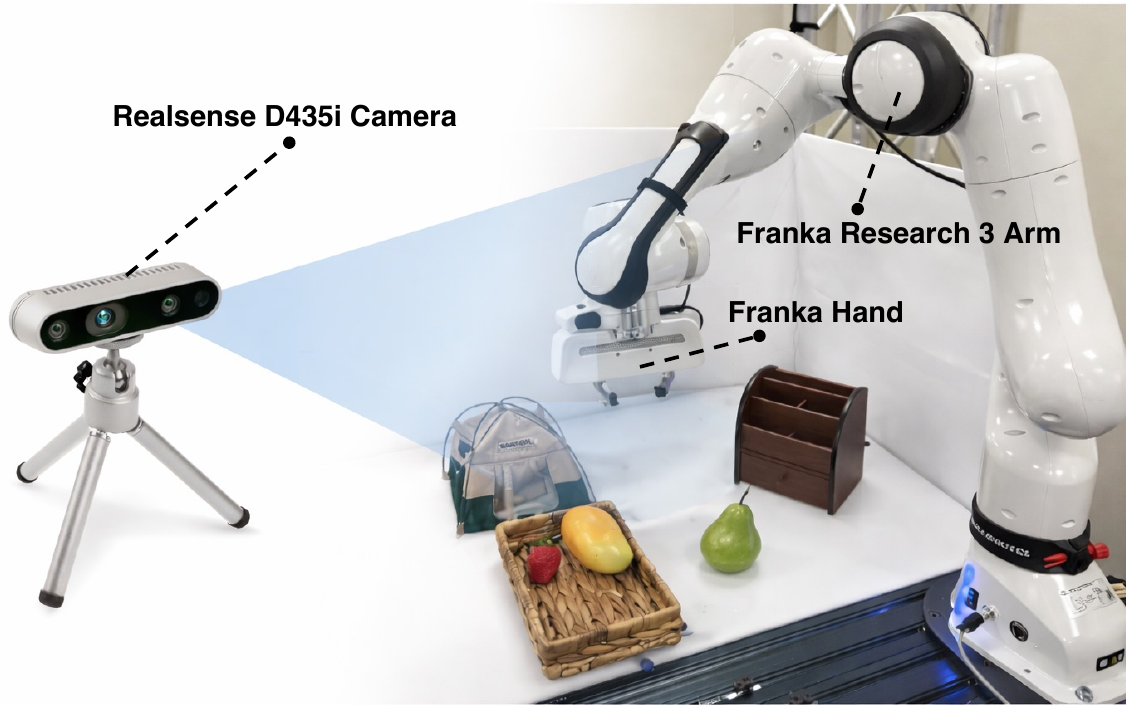} 
  \caption{\textbf{Real-world experimental workspace.} A Franka Research 3 arm with a Franka Hand gripper, and a fixed-view Intel RealSense D435i RGB-D camera are equipped for real-world evaluation.
  }
  \label{fig:workspace}
\end{figure}

\section{Experiment}

\subsection{Experiment Setup}
\label{sec:exp_setup}

We evaluate Dream2Fix from two aspects: (i) VLM-based failure understanding and structured recovery prediction, and (ii) real-world closed-loop correction that grounds the predicted recovery into robot control to improve downstream robotic manipulation performance.

\textbf{Benchmarks.}
We evaluate the VLMs on two test sets. The Dream2Fix test set is a split from our generated dataset, containing automatically generated failure records with structured labels for failure reasoning and recovery directives. The real-world failure test set is collected on the Franka Pandas robotic arm with a Realsense D435i camera and a Franka Hand (as shown in Fig~\ref{fig:workspace}), consisting of 100 trajectories across 10 manipulation tasks.

\begin{table}[t]
\centering
\caption{\textbf{Generated data quality results.} We compare verifier metrics for ground-truth demonstrations (GT) and generated failure cases, where closer values indicate more similar distributions and higher-quality generated data.}
\label{tab:dataset_distribution}
\small
\setlength{\tabcolsep}{7pt}

\resizebox{0.7\columnwidth}{!}{%
\begin{tabular}{l c c}
\toprule
Metric & GT & Generated \\
\midrule
$S_{\text{smooth}}$ & 0.814 & 0.823 \\
$S_{\text{vis}}$    & 0.999 & 0.998 \\
$S_{\text{topo}}$   & 0.977 & 0.976 \\
$S_{\text{global}}$ & 0.689 & 0.692 \\
\midrule
$\mathrm{MAE}_{xyz}(\widehat{\Delta s},\Delta s)$ & 0.008 & 0.015 \\
$\mathrm{MAE}_{rpy}(\widehat{\Delta s},\Delta s)$ & 0.034 & 0.064 \\
\midrule
$\omega$ exceedance @p95 & 0.114 & 0.222 \\
$\alpha$ exceedance @p95 & 0.339 & 0.462 \\
\bottomrule
\end{tabular}%
}
\vspace{-0.5cm}
\end{table}

\textbf{VLM Evaluation.}
We benchmark diverse state-of-the-art VLMs under a unified prompt and a structured output format. Each model takes the task instruction and an execution clip as input, and predicts (a) the success/failure result, (b) the failure stage and type when failed, and (c) a structured correction. Metrics are defined below.

\textbf{Metrics.}
We report six metrics that align with our evaluation scripts and tables:
(i) \emph{ROUGE-L} measures string-level overlap between predictions and references (for both the structured fields and the natural-language summary);
(ii) \emph{Cosine Similarity} measures semantic similarity using sentence embeddings;
(iii) \emph{Binary Success Accuracy} evaluates success/failure detection accuracy;
(iv) \emph{LLM-based Fuzzy Match} evaluates semantic correctness beyond surface-form overlap; we use Qwen3-30B-Instruct as the judge model and map its ratings (\text{correct}, \text{partially\_correct}, \text{incorrect}) to scores $(1.0,\,0.5,\,0.0)$;
(v) \emph{Correction Accuracy} assigns partial credit to near-miss recovery directives on failure samples.

\begin{figure*}[t]
  \centering
  \includegraphics[width=0.85\linewidth]{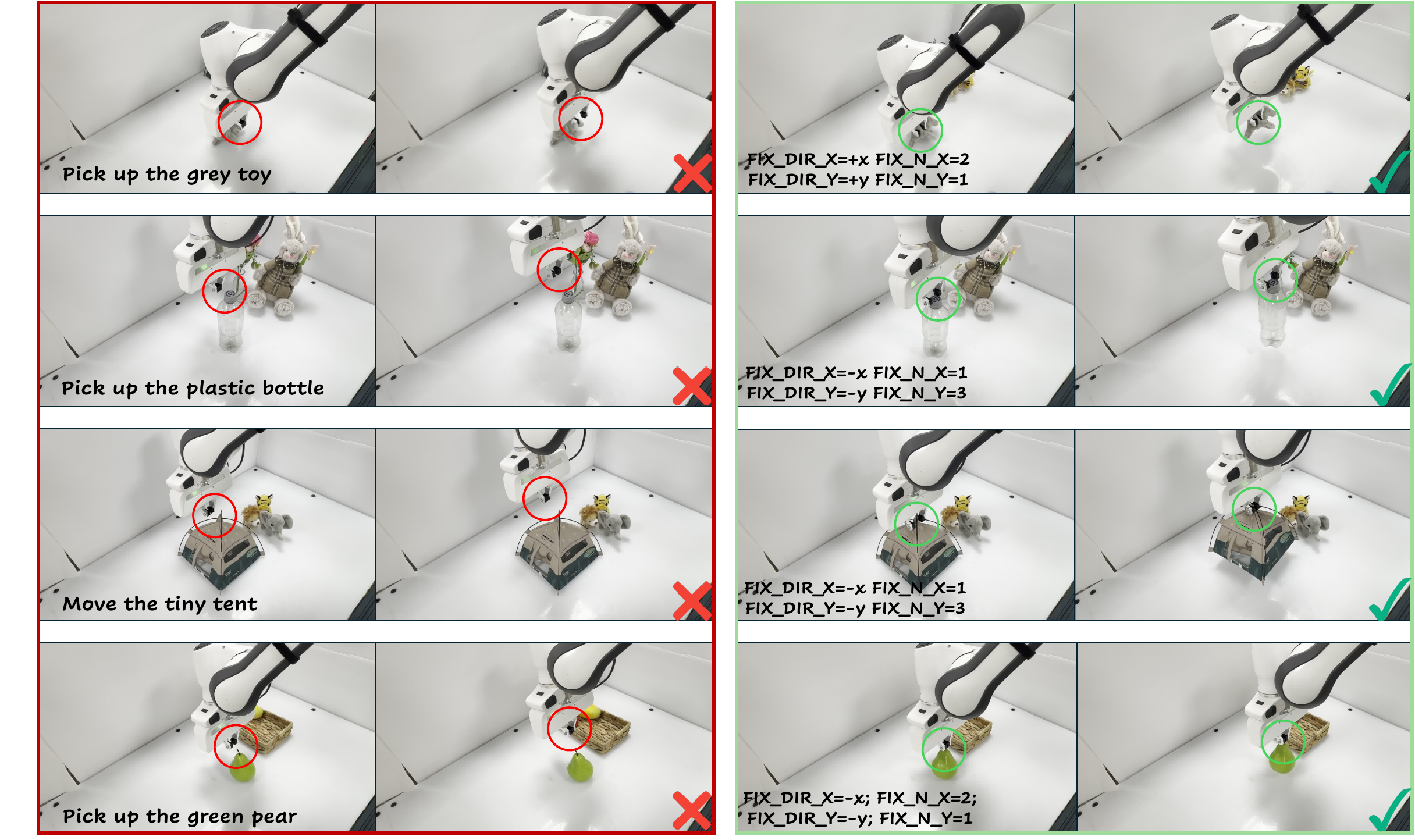} 
  \caption{\textbf{Real world robot execution.} For each task, we show the initial failed execution (left) and the corrected execution (right). Across diverse tasks, the correction adjusts the action to recover from the failure under the same real-robot setup. 
  }
  \label{fig:realworld}
\end{figure*}

For translation failures, we define correction accuracy as:
\begin{gather}
\text{Acc}=\frac{s_{\text{type}}+s_{\text{stage}}+s_x+s_y}{4} \label{eq:acc_trans}\\
s_x=\mathbbm{1}(\mathrm{dir}^{gt}_x=\mathrm{dir}^{pred}_x)\cdot
\max(0,\,1-\tfrac{|n^{gt}_x-n^{pred}_x|}{\mathrm{cap}}) \label{eq:sx}\\
s_y=\mathbbm{1}(\mathrm{dir}^{gt}_y=\mathrm{dir}^{pred}_y)\cdot
\max(0,\,1-\tfrac{|n^{gt}_y-n^{pred}_y|}{\mathrm{cap}}) \label{eq:sy}
\end{gather}
We represent the predicted translation correction along each axis $u\in\{x,y\}$ by a direction label $\mathrm{dir}^{pred}_u\in\{+,-\}$ and a magnitude level $n^{pred}_u$. Likewise, $\mathrm{dir}^{gt}_u$ and $n^{gt}_u$ are the ground-truth direction and magnitude level. The magnitude level $n_u$ is a discretized bin index of the translation offset along axis $u$ (larger $n_u$ indicates a larger offset), and $\mathrm{cap}$ denotes the maximum bin difference tolerated when computing the accuracy. Here we set $\mathrm{cap}=3$ to tolerate up to three-bin deviation in the predicted translation magnitude.

For gripper failures, we define correction accuracy as:
\begin{gather}
\text{Acc}=\frac{s_{\text{type}}+s_{\text{stage}}+s_k}{3} \label{eq:acc_grip}\\
s_k=\mathbbm{1}\!\left[|k^{gt}-k^{pred}|\le \delta_k\right] \label{eq:sk}
\end{gather}

\textbf{Real-World Experiment.}
We evaluate real-world performance in two complementary settings. (i) offline VLM evaluation: we run VLM inference on the real-world failure test set, and report failure diagnosis quality and correction accuracy. (ii) real robot evaluation: we assess downstream performance in closed-loop robot execution in two scenarios. First, we execute the predicted corrections on a physical robot and report recovery success after applying the correction. Second, we deploy OpenVLA for manipulation and use our Dream2Fix-VLM to detect failures during execution and output real-time corrections when OpenVLA fails; we report the resulting end-to-end recovery rate.

\begin{table*}[h]
\caption{\textbf{Results on the Dream2Fix test set and real-world benchmarks.} We compare baselines with Dream2Fix-VLM (ours) under the same evaluation protocol on two settings: (i) Dream2Fix (Test Set) and (ii) real-world benchmark (Zero-Shot). Metrics include text similarity (ROUGE-L and cosine similarity), match-based success measures (BinSucc and fuzzy match), and correction accuracy (Acc.). $\uparrow$ indicates higher value is better.}
\vspace{-1.0em}
\label{tab:vlm_eval}
\begin{center}
\resizebox{1\textwidth}{!}{
\begin{tabular}{lcccccccccc}
\toprule
\multirow{2}{*}{\bf Models} &
\multicolumn{5}{c}{\bf Dream2Fix Dataset (Test Set)} &
\multicolumn{5}{c}{\bf Real-World Benchmark (Zero-Shot)} \\
\cmidrule(r){2-6}\cmidrule(l){7-11}
& ROUGE$_L \uparrow$ & Cos. Sim. $\uparrow$ & BinSucc(\%) $\uparrow$ & Fuzzy Match $\uparrow$ & Acc. $\uparrow$
& ROUGE$_L \uparrow$ & Cos. Sim. $\uparrow$ & BinSucc(\%) $\uparrow$ & Fuzzy Match $\uparrow$ & Acc. $\uparrow$ \\
\midrule
LLaVA-NeXT-7B~\cite{li2024llavanext-strong}      & 0.080  & 0.099 & 0.111 & 0.322 & 0.022 & 0.000 & 0.000 & 0.000 & 0.354 & 0.000 \\
LLaVA-NeXT-34B~\cite{li2024llavanext-strong}  & 0.130   & 0.127   & 0.153   & 0.453   & 0.044    & 0.090 & 0.090 & 0.300 & 0.128 & 0.022 \\
Qwen2.5-VL-7B~\cite{bai2025qwen25vltechnicalreport}  & 0.170 & 0.391 & 0.551 & 0.241 & 0.113 & 0.052 & 0.176 & 0.250 & 0.105 & 0.008 \\
Qwen3-VL-8B~\cite{qwen3}      & 0.248 & 0.386 & 0.567 & 0.248 & 0.143 & 0.061 & 0.478 & 0.930 & 0.167 & 0.183 \\
Gemini-3-Flash~\cite{googledeepmind_gemini3_models_2025}  & 0.406 & 0.518 & 0.679 & 0.496   & 0.179 & 0.467 & 0.589 & \textbf{0.980} & 0.250 & 0.374 \\
GPT-4o                        & 0.502 & 0.538 & 0.623 & 0.615   & 0.197 & 0.190 & 0.473 & 0.720 & 0.221 & 0.126 \\
\midrule
Dream2Fix-VLM (Ours)              & \textbf{0.913} & \textbf{0.941} & \textbf{0.950} & \textbf{0.771} & \textbf{0.813}
                              & \textbf{0.621} & \textbf{0.668} & 0.820 & \textbf{0.421} & \textbf{0.472} \\
\bottomrule
\end{tabular}
}
\end{center}
\vspace{-1.5em}
\end{table*}

\subsection{Data Quality Verification.}
Across our full validation mechanism, we remove 41.9\% of synthesized rollouts and \textbf{retain 58.1\%}, resulting in the final 120+k paired dataset used in our experiments.
To characterize the distributional quality of synthesized failure cases, we compare verifier metrics between successful demonstrations and synthesized failures (as shown in Tab.~\ref{tab:dataset_distribution}). Here $s_t$ denotes the end-effector state, and $\Delta s_{t,d}=s_{t+d}-s_t$ denotes the corresponding end-effector state difference over an interval of length $d$.

For $S_{\text{smooth}}$, $S_{\text{vis}}$, $S_{\text{stop}}$, and $S_{\text{global}}$, we compute a sequence-level value for each sample and report the dataset mean. We also report the IDM consistency errors $\mathrm{MAE}_{xyz}(\Delta\hat{s},\Delta s)$ and $\mathrm{MAE}_{rpy}(\Delta\hat{s},\Delta s)$, where the IDM predicts an end-effector state difference $\Delta\hat{s}_{t,d}=f_\theta(I_t,I_{t+d})$ from two frames, and the reference difference $\Delta s_{t,d}$ is computed from recorded end-effector states used for data generation.

Because the IDM is a learned model, we first run it on successful demonstrations to obtain a reference error distribution, which reflects the prediction error of the IDM when the visual transition is real. We then apply the same IDM to synthesized failures and compute the same discrepancies between $\Delta\hat{s}_{t,d}$ inferred from the generated frames and the conditioning signal $\Delta s_{t,d}$. Comparing the resulting MAEs between successful demonstrations and synthesized failures isolates the inconsistency introduced by data generation beyond the IDM's error, and serves as a proxy for the physical and kinematic plausibility of the generated failures.

We report joint pose exceedance rates $\omega$ exceedance @p95 and $\alpha$ exceedance @p95 from the joint pose verifier. $\omega$ exceedance @p95 denotes the fraction of synthesized failures in which joint velocity exceeds a p95-calibrated velocity threshold, and $\alpha$ exceedance @p95 denotes the fraction in which joint acceleration exceeds a p95-calibrated acceleration threshold. The p95 velocity threshold $\tau_v$ and p95 acceleration threshold $\tau_a$ are calibrated on successful demonstrations by pooling $|\dot{q}_{t,j}|$ and $|\ddot{q}_{t,j}|$ over all time steps and joints, then setting $\tau_v$ and $\tau_a$ to the 95th percentile of the pooled distributions. A synthesized failure case contributes to $\omega$ exceedance @p95 if any joint at any time step exceeds $\tau_v$, and it contributes to $\alpha$ exceedance @p95 if any joint at any time step exceeds $\tau_a$.

Overall, synthesized failures remain close to successful demonstrations under our verifier metrics. The consistency scores ($S_{\text{smooth}}$, $S_{\text{vis}}$, $S_{\text{stop}}$, and $S_{\text{global}}$) stay in a comparable range across the two sets, suggesting that the synthesized failures preserve temporal smoothness, visual consistency, and coarse motion patterns observed in real executions. The IDM discrepancy measures and the $\omega$ exceedance @p95 and $\alpha$ exceedance @p95 rates fall within a moderate and controlled range, indicating that the synthesized failures maintain plausible end-effector transitions and joint motions relative to successful demonstrations. These results suggest that Dream2Fix produces failure cases with good visual fidelity and physical plausibility under our verifier metrics, supporting their use for learning and evaluating recovery.

\vspace{-0.1cm}

\subsection{Quantitative Results}
We evaluate Dream2Fix in two settings: the Dream2Fix benchmark (test set) and a real-world benchmark under zero-shot manner. Table~\ref{tab:vlm_eval} summarizes results for diverse baselines, including open-source VLMs (LLaVA-NeXT-7B~\cite{li2024llavanext-strong}, LLaVA-NeXT-34B~\cite{li2024llavanext-strong}, Qwen2.5-VL-7B~\cite{bai2025qwen25vltechnicalreport}, and Qwen3-VL-8B~\cite{qwen3}) and closed-source models (Gemini-3-Flash and GPT-4o). To make the comparison as fair as possible, all models are prompted with the same structured output format that matches our annotation schema.

On the Dream2Fix benchmark, Dream2Fix-VLM achieves the best overall performance and delivers a substantial gain in correction accuracy. Specifically, Acc improves from 0.197 (GPT-4o) to 0.813, showing that Dream2Fix-VLM produces much more accurate recovery outputs rather than only identifying failures. Dream2Fix-VLM also attains the highest binary success rate, which is consistent with stronger recovery decisions beyond failure recognition.

For the real-world benchmark, Dream2Fix-VLM is evaluated in zero-shot, without fine-tuning on real-world data. Metrics are lower than on the Dream2Fix benchmark, since real-world data contains more varied appearances and execution conditions. Notably, several baselines achieve high binary success rates but still obtain low correction accuracy, suggesting that they can often judge success or failure while failing to produce correct recovery outputs. In contrast, Dream2Fix-VLM delivers the strongest correction accuracy on real-world data (Acc 0.472), supporting improved failure reasoning and correction capability under strict zero-shot evaluation, and demonstrating the usefulness of Dream2Fix data for learning corrective recovery in real-world settings.

\subsection{Downstream Robotic Experiments}
We evaluate downstream performance in real-world execution (as shown in Fig.~\ref{fig:realworld}) using two complementary settings.
First, we test whether our VLM-predicted corrections can be executed on a real robot to recover from failures in the Real-World Benchmark. These cases reflect failures observed in real data, and the robot executes the predicted trajectory correction to attempt recovery. Second, we test recovery during deployment with OpenVLA~\cite{kimopenvla}. We run OpenVLA on a set of manipulation tasks, collect its deployment failures, and then apply our predicted corrections to recover from these failures.
In both settings, the Dream2Fix-VLM is trained only on Dream2Fix synthetic data and is applied to real-world data in a strict zero-shot manner. For each setting, we conduct 50 runs and report the recovery rate, computed over failure cases, where a failure is counted as recovered if executing the predicted correction leads to task success. Tab.~\ref{tab:real_world_recovery} summarizes the recovery rates for Real-World Benchmark and OpenVLA deployment failures.

\begin{table}[t]
\centering
\caption{\textbf{Real-world results.} We report recovery rate ($\uparrow$ is better) for two methods, Dream2Fix-VLM and Gemini3-Flash, evaluated on failures from our Real-World Benchmark and OpenVLA deployment.}
\label{tab:real_world_recovery}
\small
\setlength{\tabcolsep}{6pt}
\resizebox{0.8\columnwidth}{!}{%
\begin{tabular}{l cc}
\toprule
\multirow{2}{*}{Failure Source} & \multicolumn{2}{c}{Recovery Rate (\%) $\uparrow$} \\
\cmidrule(lr){2-3}
& Dream2Fix-VLM & Gemini3-Flash \\
\midrule
Real-World Benchmark & 46\% & 36\% \\
OpenVLA Deployment     & 40\% & 30\% \\
\bottomrule
\end{tabular}%
}
\vspace{-0.5cm}
\end{table}

\section{Conclusions}
We presented Dream2Fix, a data generation pipeline that synthesizes photorealistic manipulation failures from successful demonstrations and curates them with visual and physical verifier criteria. Using the resulting data, we trained Dream2Fix-VLM to produce actionable trajectory corrections beyond failure judgment. Across our validation mechanism, we remove 41.9\% of synthesized rollouts and retain 58.1\%, resulting in a final dataset of over 120k paired failure–correction samples. This retention ratio highlights that, while the current action-conditioned world model already serves as a strong controllable simulator for diverse failure generation, a substantial fraction of rollouts still deviates from realistic visual and physical behavior, which leaves room to further improve physical plausibility. Experiments on the Dream2Fix benchmark and real-world data show strong zero-shot correction accuracy, improving from 19.7\% to 81.3\%, and real-robot trials confirm that the predicted corrections can be executed to recover from failures. Future work will improve the action-conditioned dynamics of our robotic world model to produce more physically consistent rollouts, enabling higher-fidelity data augmentation at scale.

\addtolength{\textheight}{-12cm}   




\bibliographystyle{ieeetr}
\bibliography{my_bib}

\end{document}